\documentclass{article}
\usepackage{ijcai22}

\usepackage{xr-hyper}

\usepackage{times}

\usepackage{soul}
\usepackage{url}
\usepackage[hidelinks]{hyperref}
\usepackage[utf8]{inputenc}
\usepackage[small]{caption}
\usepackage{graphicx}
\usepackage{amsmath}
\usepackage{booktabs}
\usepackage{xcolor}

\usepackage{amsfonts}

\title{Benchmarking Perturbation-based Saliency Maps for Explaining Atari Agents}

\author{blinded for review}

\author{
Tobias Huber$^1$\footnote{Contact Author}\and
Benedikt Limmer$^1$\and
Elisabeth André$^1$\\
\affiliations
$^1$University of Augsburg\\
}

\newcommand{\citet}[1]{\citeauthor{#1}~\shortcite{#1}}
\newcommand{\citep}{\cite}

\newcommand{\occlusion}{Occlusion Sensitivity}

\begin{document}

\maketitle

\begin{abstract}
One of the most prominent methods for explaining the behavior of Deep Reinforcement Learning (DRL) agents is the generation of saliency maps that show how much each pixel attributed to the agents' decision. However, there is no work that computationally evaluates and compares the fidelity of different saliency map approaches specifically for DRL agents. It is particularly challenging to computationally evaluate saliency maps for DRL agents since their decisions are part of an overarching policy. For instance, the output neurons of value-based DRL algorithms encode both the value of the current state as well as the value of doing each action in this state. This ambiguity should be considered when evaluating saliency maps for such agents. In this paper, we compare five popular perturbation-based approaches to create saliency maps for DRL agents trained on four different Atari 2600 games. The approaches are compared using two computational metrics: dependence on the learned parameters of the agent (sanity checks) and fidelity to the agent's reasoning (input degradation). During the sanity checks, we encounter issues with one approach and propose a solution to fix these issues. For fidelity, we identify two main factors that influence which saliency approach should be chosen in which situation.

\end{abstract}

\section{Introduction}
\label{sec:introduction}

With the rapid development of machine learning methods, Deep Reinforcement Learning (DRL) agents are making their way into increasingly high-risk applications, such as healthcare and robotics.
However, this comes with an increasing complexity of state spaces and algorithms, making it hardly if at all possible to comprehend the decisions of these agents \citep{Heuillet2021}.
The research areas of Explainable Artificial Intelligence (XAI) and Interpretable Machine Learning aim to shed light on the decision-making process of such black-box models.
In the case of DRL agents, which utilize neural networks with visual inputs, the most common explanation approach is the generation of saliency maps that highlight the most relevant input pixels for a given decision. 
In general, there are three main ideas on how to create saliency maps.
The first idea is to use the gradient with respect to each input to see how much small changes of this input influence the prediction.
The second group of methods uses modified propagation rules to calculate how relevant each neuron of the network was, based on the intermediate results of the prediction.
Finally, perturbation-based approaches perturb areas of the input and measure how much this changes the output of the network.
Both gradient and modified propagation saliency maps have been applied to DRL agents \citep{zahavy2016,huber2019}.
However, recent years saw a trend towards perturbation-based saliency maps \citep{greydanus2018visualizing,puri2020}.
The major advantage of perturbation-based approaches is their model agnosticism. 
Since they only use the in- and outputs of the agent, they can be applied to any agent without adjustments.

If saliency maps are used to analyze DRL agents in high-risk applications, it is crucial that we can rely on the information provided by the saliency map.
That is, the most relevant pixels should actually be the most relevant input regions for the agent's strategy.
This is often called fidelity of an explanation technique \citep{mohseni2020multidisciplinary}.
The need for evaluating the fidelity of saliency maps was further demonstrated by \citet{adebayo2018sanity}.
They proposed sanity checks which showed that for some saliency approaches, there is no strong dependence between the learned parameters of image classifiers and the saliency maps that analyze their underlying neural network.
Surprisingly, there are no computational evaluations that assess and compare the fidelity of different saliency maps for DRL agents.
This is despite the fact that DRL agents are more challenging to analyze than classification models \citep{Heuillet2021}.
The decisions of a DRL agent are not isolated but are part of an overarching policy and might be influenced by delayed rewards, which may not be discernible in the current state.
This makes it even more challenging to verify whether a saliency map matches the internal reasoning of a DRL agent.
In the prominent family of value-based DRL algorithms, for example, the output values do not only describe the probability of choosing an action.
They also encode the estimated value of the input state for the current policy.
This ambiguity is often ignored when saliency maps are applied to analyze the decisions of value-based DRL agents.

In this paper, we present, to the best of our knowledge, the first computational fidelity evaluation of different saliency maps for DRL agents.
In particular, we make the following contributions.
By focusing on five perturbation-based saliency approaches, this work gives an overview of which approaches should be used in what situation by practitioners who don't have full access to their DRL agent's model.
One drawback of perturbation-based saliency maps is that they depend on a choice of parameters.
To ensure that all of the algorithms tested in this paper perform reasonably well, we present a novel methodology to fine-tune the parameters of perturbation-based saliency maps for DRL agents. 
Furthermore, we propose a way to separately measure how well a saliency map captures an agent's respective action- and state-value estimation.
We demonstrate that the performance of saliency map approaches differs considerably when measuring state-values compared to action-values.

As test-bed for our evaluation, we use the Atari 2600 environment.
As metrics, we use the sanity checks proposed by \citet{adebayo2018sanity} and an insertion metric that measures if the most relevant pixels, according to the saliency map, actually influence the agent's decision.
As far as we know, this is the first time that sanity checks are done for different perturbation-based saliency maps for any kind of model.
\section{Related Work}
\label{sec:related_work}

In general, evaluation metrics for XAI approaches can be separated into two broad categories:
human user studies and computational measurements \citep{mohseni2020multidisciplinary}.
So far, DRL agents are mostly evaluated with user studies.
\citet{huber2020LocalandGlobal} and \citet{anderson2019mere-mortals} conduct user studies to evaluate a single variant of modified propagation and perturbation-based saliency maps respectively, with regards to mental models, trust, and user satisfaction. 
\citet{puri2020} investigate whether perturbation-based saliency maps can help participants with chess puzzles, by highlighting which pieces were relevant for an agent's solution for these puzzles. 
\citet{greydanus2018visualizing} test whether participants can identify overfit policies with the help of perturbation-based saliency maps.
However, exclusively relying on user studies might only measure how convincing the saliency maps look but not how much they reflect the agent's internal reasoning.
Therefore, it is important to additionally evaluate the fidelity of saliency maps through computational measurements \citep{mohseni2020multidisciplinary}.
Such measurements also provide an easy way to collect preliminary data before recruiting users for a user study. 

There is a growing body of work on computationally evaluating the fidelity of saliency maps for image classification models.
The most common measurement is \emph{input degradation}.
Here, the input of the model is gradually perturbed, starting with the most relevant input features according to the saliency map.
For visual input, this is either done by perturbing individual pixels per step \citep{petsiuk2018rise,ancona2018better_understanding_gradient} or by perturbing patches of the image in each step
\citep{samek2017evaluation_visualization,kindermans2018pattern_attribution,schulz2020restricting_the_flow}.
If the saliency map matches the model's reasoning, then the model's confidence should fall quickly.
In addition to perturbing features, some newer approaches also propose an insertion metric where they start with fully perturbed inputs and gradually insert relevant features \citep{ancona2018better_understanding_gradient,petsiuk2018rise,schulz2020restricting_the_flow}.
Recently, \citet{Tomsett20} demonstrated that input degradation can be unreliable and is sensitive to implementation details like the type of perturbation.
They conclude that researchers should employ several versions of this metric and try to understand potential reasons for unreliability. 

Another prominent computational measurement for saliency maps for image classification models are the so-called \emph{sanity checks} proposed by \citet{adebayo2018sanity}.
These tests measure whether the saliency maps are dependent on what the model's neural network learned. 
One method for this is gradually randomizing the layers of the neural network and measuring how much this changes the saliency maps.
If the saliency maps are  faithful to what the network learned then they should change considerably for each randomized layer. 
\citeauthor{adebayo2018sanity} did this for various gradient-based approaches and \citet{sixt2019} additionally tested modified propagation methods.
Both groups found that some approaches did not really depend on the parameters of the network and therefore can't faithfully reflect the model's internal reasoning.
As far as we know, there is no work that verified whether different perturbation-based saliency maps depend on the network's learned parameters even though this is one of the most popular saliency map approaches.

For DRL agents, there is very little work on computationally evaluating the fidelity of saliency maps.
\citet{puri2020} recorded which chess pieces human experts identified as important in a set of chess puzzles.
This allows them to computationally compare these pieces to the pieces that saliency maps identify as relevant for an agent. 
However, this does not measure the saliency maps' fidelity to the agent's reasoning, but whether the saliency maps coincide with human reasoning.
\citet{huber2020LocalandGlobal} calculate sanity checks for a single modified propagation saliency approach.
\citet{Atrey2020Exploratory} conduct experiments to verify hypotheses that are generated from observing saliency maps.
However, both the formulation of hypotheses as well as their verification rely on manual inspection of the saliency maps.
In this sense, we see our paper as the first computational evaluation to benchmark the fidelity of different saliency map approaches for DRL agents.

\section{Experiments}
\label{sec:experiments}

\textbf{The test-bed} in our paper is the Atari Learning Environment \citep{bellemare2013ALE}.
Four DRL agents were trained on the games MsPacman (simplified to Pac-Man in this work), Space Invaders, Frostbite, and Breakout using the Deep Q-Network (DQN) \citep{mnih2015human} implementation of the OpenAI Baselines Framework \citep{baselines}.
We chose the DQN because it is the most basic DRL architecture which most other DRL agents build upon. 
The games were selected because the DQN performs very well on Breakout and Space Invaders but performs badly on Frostbite and Pac-Man.
We slightly adjusted the reward function compared to \citet{mnih2015human} to enhance the performance of our agents.
The reward is given by the change in in-game score since the last state, which we scaled such that the minimal possible reward is 1.
All experiments were done on the same machine with an Nvidia GeForce GTX TITAN X GPU to ensure comparability of the results.
Our code is available online.\footnote{ 
\url{https://github.com/belimmer/PerturbationSaliencyEvaluation}}

As \textbf{saliency map approaches}, we chose \occlusion{} \citep{zeiler2014visualizing} since it is the first and most basic perturbation-based saliency approach.
Furthermore, we use LIME \citep{ribeiro2016should} and RISE \citep{petsiuk2018rise} which are two of the most popular perturbation-based saliency maps in general.
Finally, we chose two approaches that were specifically proposed for DRL: Noise Sensitivity \citep{greydanus2018visualizing} and SARFA \citep{puri2020}.
Detailed descriptions of the different approaches can be found in Appendix \ref{sec:saliency_maps}.
We evaluate the generated saliency maps using two different computational metrics: Sanity checks and an insertion metric.

The \textbf{sanity checks} proposed by \citet{adebayo2018sanity} measure the dependence between the saliency maps and the parameters learned by the neural network of the agent. 
To this end, the parameters of each layer in the network are randomized in a cascading manner, starting with the output layer.
Every time a new layer is randomized, a saliency map for this version of the agent is created. 
The resulting saliency maps are then compared to the saliency map for the original network, using three different similarity metrics (Spearman rank correlation, Structural Similarity (SSIM), and Pearson correlation of the Histogram of Oriented Gradients (HOGs)).
Following \citet{sixt2019}, we account for saliency maps that differ only in sign by additionally computing similarity with an inverse version of the saliency maps and using the maximum similarity.
We tuned the similarity metrics analogous to \citet{adebayo2018sanity} (see Appendix \ref{ap:similarity_metric_test}).
If the saliency maps depend on the learned parameters of the agent then the saliency maps for the randomized model should vastly differ from the ones of the original model.
For our tests, we calculate the sanity checks for $1000$ states of each game.

\begin{figure}
    \centering
    \includegraphics[width=0.8\linewidth]{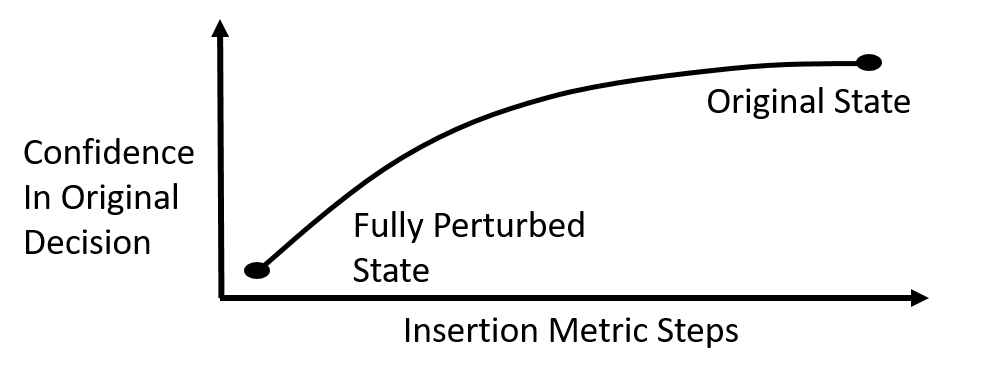}
    \caption{A schematic representation of the insertion metric curve.}
    \label{fig:insertion_metric_sheme}
\end{figure}

If a saliency map is faithful to the agent, then the most relevant pixels should have the highest impact on the agent's decision.
To test this property, we use a \textbf{insertion metric} similar to \citet{petsiuk2018rise}.
We do not use a deletion metric, since we feel that it is too similar to the way that perturbation-based saliency maps are created.
The insertion metric starts with a fully perturbed state.
In each step, $84$ perturbed pixels (approx. $1.2\%$ of the full state) are uncovered, starting with the most relevant pixels according to the saliency map.
For LIME, the superpixels are sorted by their relevance but the order of pixels within superpixels is randomized.
The partly uncovered state is then fed to the agent and its output for the original action, which the agent chooses in the unperturbed state, is measured.
If the saliency map correctly highlights the most important pixels, then the agent's confidence should increase quickly for each partly uncovered image.
Plotting the agent's confidence in each step of the insertion metric results in an insertion metric curve (Fig. \ref{fig:insertion_metric_sheme}). 
If the confidence increases quickly, then the area under the insertion curve is high. 
Therefore, the Area Under the insertion metric Curve (AUC) is used to represent the result of the insertion metric for a single state.
Before we can apply the insertion metric to our DRL agents, we have to decide how to perturb the input and which output value we measure in each step. 

\citet{Tomsett20} found that the choice of perturbation has a high impact on the result of the metric.
To be more robust against this influence, we use two different perturbations: black occlusion and uniform random perturbation.
Black is similar to the background color in most Atari games and therefore acts as ``deleting'' features from the state.
Uniform random perturbation performed well for \citeauthor{Tomsett20}.

Next, we have to decide which output we want to measure. 
This comes with two challenges.
First, the output q-values of value-based reinforcement learning algorithms like the DQN do not directly describe the agent's confidence in particular actions.
Instead, they approximate the value of the current state in combination with each action.
To disentangle this ambiguity, we propose to use two different sub-metrics.
One measures how well the saliency map identifies features relevant to the state-value, and the other measures the same for the action-value.
For the action-value, we propose to use the advantage as defined by \citet{wang16dueling}.
For the state-value, we suggest using the q-value.
The second challenge is that a reliable metric should not be distorted by outliers.
For our Pac-Man agent, for example, we observed states with q-values around $1$ and other states with q-vales around $50$.
To reduce the effect of outlier states, we tested several methods of normalizing the agent's output during the insertion metric.
To this end, we used $28$ different variants of \occlusion{} saliency maps.
For each variant, we calculated differently normalized insertion metrics over $1000$ states of the Pac-Man environment.
\citet{Tomsett20} suggest using a low Standard Deviation (SD) as an early indicator for reliable saliency map metrics.
Therefore, we chose the normalization method that resulted in the lowest SD of the area under the insertion curve across the $1000$ states. 
For the advantage, we obtained the lowest SD if we did not use any normalization.
The q-values got the lowest SD when we divided each insertion metric step by the result of the original state.
For more details about the normalization tests see Appendix \ref{ap:normalizing_insertion_metric}. 

For our final evaluation of the different saliency methods, we use $1000$ states of each of the four Atari games.
For each of those states, we calculated the insertion metric in four different variants: measuring the advantage of the chosen action with random and black perturbation, and measuring the normalized q-value with random and black perturbation.

\section{Parameter Tuning}
\label{sec:parameter_tuning}

One of the biggest drawbacks of perturbation-based saliency map approaches is that they depend on a choice of parameters.
Before we can run our experiments we have to find suitable parameters.
This is often done by manually tuning the parameters until the resulting saliency maps look reasonable.
However, tuning the parameters in this way does not guarantee that the saliency maps match the agent's internal reasoning.
To obtain a fidelity benchmark for saliency maps, we computationally tune the parameters to perform well in the insertion metric.
We do not tune the parameters for the sanity checks, since sanity checks do not measure how well a saliency map approach performs. 
Instead, they identify which approaches do not work at all.
To tune the parameters for our final tests we need to decide on two things: which states we test the parameters on and how we combine the results from the four different insertion metric variants.

To combine the results of the random and black insertion metric variants, we measure the mean of the area under the insertion curve over both the black and the random perturbation insertion metric.
For our evaluation, we would also like to find parameters that are able to analyze both the agent's action-value and state-value estimation.
To this end, we standardize the mean results of the aforementioned tests for the advantage and q-values measurements respectively.
The sum of these standardized values is then used as a single value that  measures the performance of the parameters.
Finally, to ensure comparability between approaches and to be able to run our final experiment in a reasonable amount of time, we did not choose the top parameters.
Instead, we use the best parameters that needed up to three seconds to compute a single saliency map.

As test set for our parameter tuning, it is not feasible to use the full stream of $1000$ states that we want to use in our final evaluation.
LIME and RISE in particular have long computation times and a large number of possible parameter combinations. 
This would make the run-time of the parameter test explode.
For more information on the run-time of each saliency approach, see Appendix \ref{ap:runtime}.
Therefore, we need to find a suitable subset of states (we used $10$ states) that represent as many states as possible.
Since there are no test- or validation-sets in reinforcement learning we have to choose these subsets from the full stream of gameplay.
As potential candidates, we tested $10$ randomly selected subsets of states and $12$ subsets selected by different variants of the HIGHLIGHT-DIV algorithm.
This algorithm selects a diverse set of states that give a good overview of the agent's policy \citep{amir2018highlights}.  
To compare how well these subsets represent the full stream of gameplay, we calculated the combined insertion metric results, as described above, for the full $1000$ states of Pac-Man using $28$ different parameter combinations of \occlusion{}. 
The particular parameters were chosen since they are fast to compute.
Based on these results we obtained a “ground truth” for how those $28$ parameters for \occlusion{} should be ranked.
Now, a subset of states is suited for searching parameters if the parameter ranking obtained by the subset is similar to the ranking obtained by the full $1000$ states.
To calculate the similarity of different rankings we used both Spearman's and Kendall rank correlation coefficients.
We found that HIGHLIGHTS-DIV only performed well when the diversity threshold was very high.
This threshold makes sure that the selected states are not too similar to each other. 
When the threshold was low the HIGHLIGHTS-DIV states performed worse than the random ones.
We got the best results when the threshold was so high that increasing the threshold resulted in subsets with less than 10 states since the algorithm could not find any more states that could be added to the subset.
For the exact correlation values and the HIGHLIGHTS-DIV variants see Appendix \ref{ap:parameter_tuning_state_selection}.

In total, we tested $4918$ parameter combinations across all three methods. 
Since LIME is very dependent on the choice of a segmentation function, we tested parameters for the three most common Segmentation techniques \emph{SLIC}, \emph{Quickshift} and \emph{Felzenszwalb}.
The final parameters that we used in our experiments, together with all the parameter ranges that we tested can be found in Appendix \ref{ap:parameter_tuning_used_parameters}. 

\begin{figure}
\centering
\newcommand{\mysize}{0.22}
\newcommand{\figsize}{0.9}
\newcommand{\myskip}{0\baselineskip} 
 \begin{minipage}[t]{\mysize\linewidth}
    \parbox[t][\myskip]{\linewidth}{\centering
        Input State}
        \includegraphics[width=\figsize\linewidth]{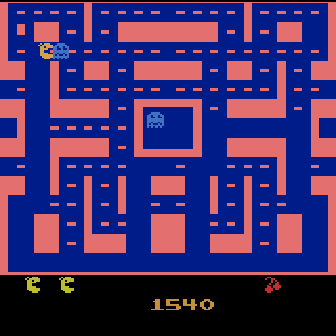}
        
    \end{minipage}
    \begin{minipage}[t]{\mysize\linewidth}
    \parbox[t][\myskip]{\linewidth}{\centering
        Occlusion} 
         \includegraphics[width=\figsize\linewidth]{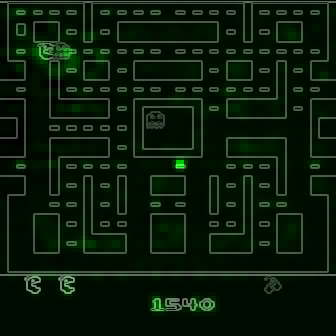}
    \end{minipage}
    \begin{minipage}[t]{\mysize\linewidth}
    \parbox[t][\myskip]{\linewidth}{\centering
        \emph{NS}}
        \includegraphics[width=\figsize\linewidth]{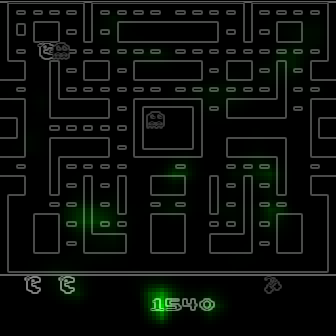}
    \end{minipage}
    \begin{minipage}[t]{\mysize\linewidth}
    \parbox[t][\myskip]{\linewidth}{\centering
        Sarfa}
        \includegraphics[width=\figsize\linewidth]{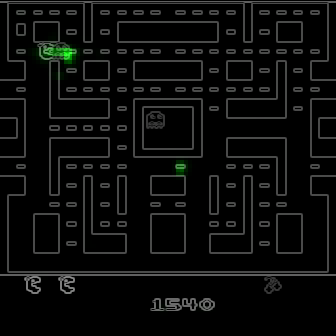}
    \end{minipage}
    \begin{minipage}[t]{\mysize\linewidth}
    \parbox[t][\myskip]{\linewidth}{\centering
        RISE}
        \includegraphics[width=\figsize\linewidth]{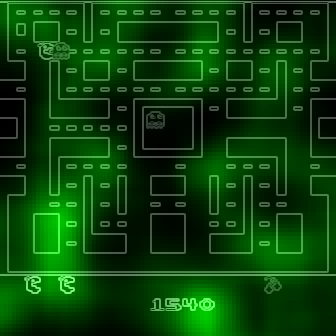}
    \end{minipage}
    \begin{minipage}[t]{\mysize\linewidth}
    \parbox[t][\myskip]{\linewidth}{\centering
        LIME Felz}
        \includegraphics[width=\figsize\linewidth]{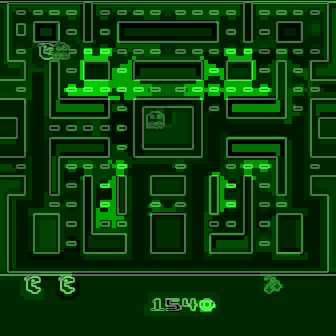}
        
    \end{minipage}
    \begin{minipage}[t]{\mysize\linewidth}
    \parbox[t][\myskip]{\linewidth}{\centering
        LIME Quick}
         \includegraphics[width=\figsize\linewidth]{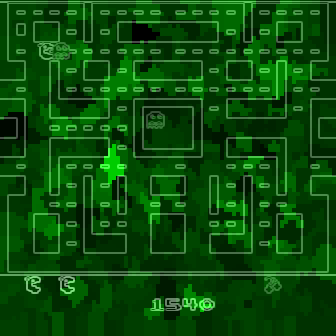}
        
    \end{minipage}
    \begin{minipage}[t]{\mysize\linewidth}
    \parbox[t][\myskip]{\linewidth}{\centering
        LIME SLIC}
         \includegraphics[width=\figsize\linewidth]{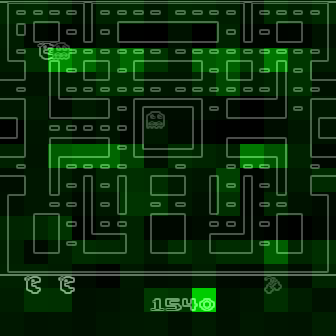}
        
    \end{minipage}
    
\caption{Example saliency maps for a Pac-Man game state generated by each of the approaches investigated in this paper (\emph{NS} is Noise Sensitivity).
}
\label{fig:expl_pac}
\end{figure}

\begin{figure}
    \centering
    \includegraphics[width=\linewidth]{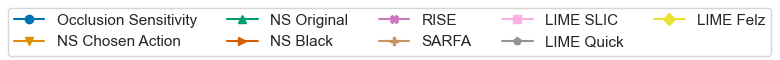}
    
    \newcommand{\myspace}{10pt} 
    \begin{minipage}{0.06\linewidth}
            \vspace{-10pt}
            1.0
            
            \vspace{6pt}
            0.6
            
            \vspace{18pt}
            0.0
    \end{minipage}
    \begin{minipage}{0.3\linewidth}
    \centering
        \includegraphics[width=\linewidth]{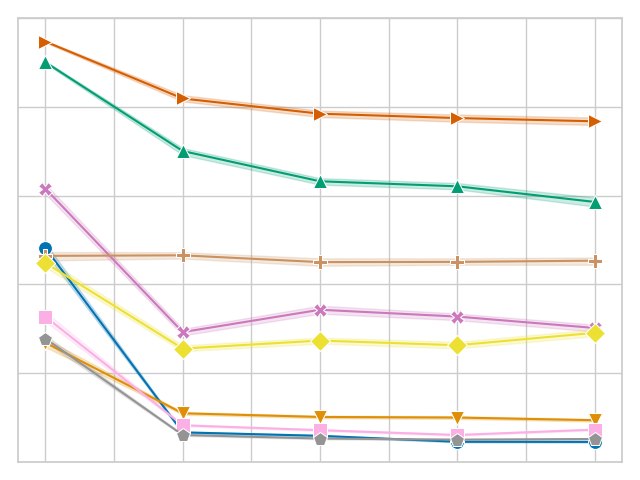}
        \parbox[t][0.1\baselineskip]{\linewidth}{\centering Spearman}
    \end{minipage}
     \begin{minipage}{0.3\linewidth}
    \centering
        \includegraphics[width=\linewidth]{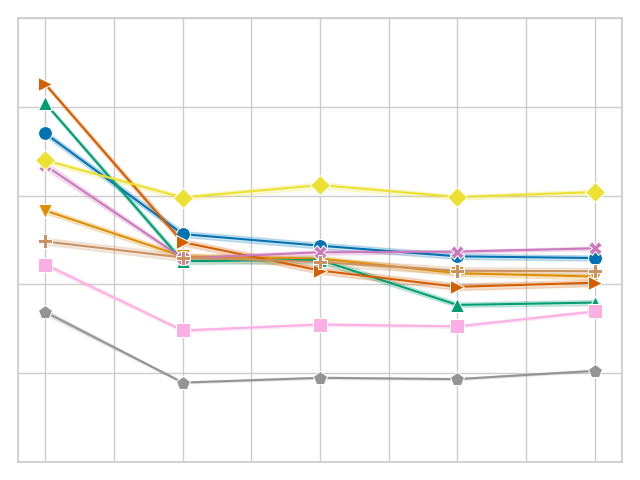}
         \parbox[t][0.1\baselineskip]{\linewidth}{\centering SSIM}
    \end{minipage}
    \begin{minipage}{0.3\linewidth}
    \centering
        \includegraphics[width=\linewidth]{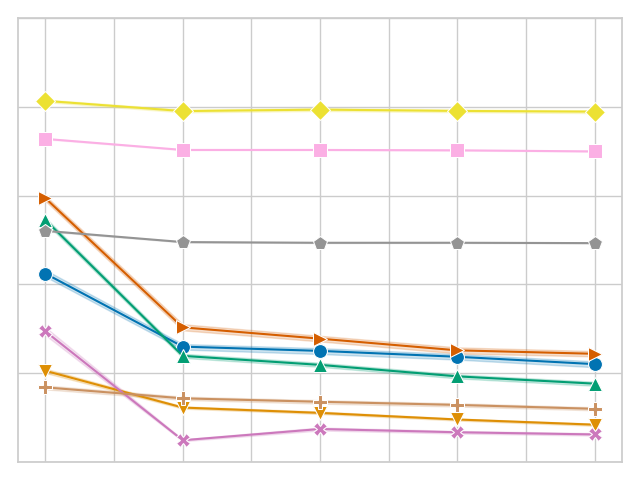}
        \parbox[t][0.1\baselineskip]{\linewidth}{\centering Pearson}
    \end{minipage}
    \caption{Results of the sanity checks for the different saliency map approaches.
    Measured for $1000$ states of each of the 4 tested games. 
    Starting from the left, each mark represents an additional randomized layer starting with the output layer.
    The y-axis shows the average similarity values (Spearman rank correlation, SSIM, Pearson correlation of the HOGs). 
    High values indicate a low parameter dependence.
    The  translucent error bands show the 99\% CI but are barely visible due to low variance in the results.}
    \label{fig:sanity_checks}
\end{figure}

\section{Results}
\label{sec:results}

Fig. \ref{fig:expl_pac} shows example saliency maps for a Pac-Man state (saliency maps for the remaining agents are shown in Appendix \ref{ap:additonal_results}).
To prevent cherry-picking, the state is chosen by the HIGHLIGHTS-DIV algorithm \citep{amir2018highlights}.

The results of the \textbf{sanity checks} test are shown in Fig. \ref{fig:sanity_checks}.
The lower the scores the higher the dependence on the agents' learned parameters.
An example for the different saliency maps during a single run of the sanity check can be seen in Appendix \ref{ap:additonal_results}.
Notably, LIME has a very high Pearson correlation of HOGs.
Furthermore, the original Noise Sensitivity has low dependence on the parameters of the output layer when compared to Occlusion Sensitivity.
Since those two approaches are very similar in theory, we implemented two modifications of Noise Sensitivity to investigate the reason for this difference in parameter dependence.
First, \emph{Noise Sensitivity Black} occludes the circles in the Noise Sensitivity approach with black color instead of blurring them.
Second, \emph{Noise Sensitivity Chosen Action} changes the way that the importance of each pixel is calculated from the original equation (Eq. \ref{sec:saliency_maps} (\ref{eq:noise})), which takes all actions into account, to the one used by Occlusion Sensitivity (Eq. \ref{sec:saliency_maps} (\ref{eq:occlusion})), which focuses on the chosen action.
We did not test a combination of black circles and the Occlusion Sensitivity importance calculation since that would be equivalent to Occlusion Sensitivity with circles instead of squares.
While the black occlusion did not really change the sanity check results, the change of the importance calculation immensely increased the dependence on the learned parameters.

\begin{table*}
\begin{tabular}{lrrrrrrrr}
\toprule
Metric &	Occlusion	& Noise &	SARFA &	RISE & LIME Felz & LIME quick & LIME slic & Baseline \\
\midrule
Pac-Man: \\
Q-val rand     &   0.54$\pm$1.3 &   0.75$\pm$0.7 &   0.76$\pm$1.2 &        \textbf{1.1$\pm$2.0} &                    0.46$\pm$0.7 &                  0.67$\pm$1.1 &                0.62$\pm$1.1 &   0.85$\pm$1.5 \\
Adv rand &  -0.52$\pm$1.2 &  -0.03$\pm$0.8 &  -0.74$\pm$1.3 &      \textbf{-0.01$\pm$1.1} &                   -0.43$\pm$1.2 &                 -0.44$\pm$1.0 &               -0.36$\pm$1.1 &  -0.22$\pm$1.0 \\
Q-val black      &   \textbf{3.08$\pm$3.2} &   0.66$\pm$0.8 &   0.83$\pm$1.8 &       1.01$\pm$1.8 &                    2.83$\pm$5.3 &                  2.49$\pm$4.7 &                2.47$\pm$4.4 &   0.53$\pm$0.8 \\
Adv black  &   1.23$\pm$1.6 &   0.15$\pm$0.3 &    \textbf{1.7$\pm$0.8} &       0.21$\pm$0.4 &                    0.64$\pm$0.7 &                  0.94$\pm$0.5 &                0.67$\pm$0.5 &   0.06$\pm$0.3 \\

Breakout: \\

Q-val rand & -0.72$\pm$2.5 &  -1.01$\pm$3.0 &  -3.19$\pm$3.9 &      -0.97$\pm$2.7 &                   -0.98$\pm$2.7 &                 \textbf{-0.48$\pm$4.1} &               -0.53$\pm$3.2 &  -2.21$\pm$2.9 \\
Adv rand & -0.42$\pm$4.7 &  -1.52$\pm$8.4 &  -0.92$\pm$8.4 &       \textbf{0.85$\pm$6.1} &                    -0.7$\pm$6.5 &                 -0.54$\pm$5.4 &               -0.05$\pm$4.8 &  -0.76$\pm$5.8 \\
Q-val black      &   3.16$\pm$4.2 &   3.04$\pm$4.2 &   1.97$\pm$2.0 &       3.39$\pm$4.2 &                    \textbf{7.48$\pm$9.6} &                   5.8$\pm$8.7 &                6.12$\pm$9.7 &   2.13$\pm$3.1 \\
Adv black  &   0.02$\pm$0.5 &   0.19$\pm$0.6 &   0.53$\pm$1.1 &       0.29$\pm$0.6 &                    0.24$\pm$0.6 &                  0.24$\pm$0.4 &                \textbf{0.71$\pm$1.4} &   0.07$\pm$0.2 \\

Frostbite: \\

Q-val rand     &  0.56$\pm$1.0 &  0.83$\pm$1.0 &  0.73$\pm$1.0 &       \textbf{0.92$\pm$1.1} &                    0.75$\pm$0.9 &                  0.37$\pm$1.0 &                0.36$\pm$1.0 &  0.88$\pm$1.1 \\
Adv rand &  0.31$\pm$1.1 &  0.38$\pm$1.2 &   0.2$\pm$1.2 &       0.35$\pm$0.9 &                     0.2$\pm$0.9 &                  0.24$\pm$1.3 &                0.23$\pm$1.3 &   \textbf{0.4$\pm$1.2} \\
Q-val black     &  \textbf{5.65$\pm$3.1} &  0.58$\pm$0.2 &  1.53$\pm$1.6 &        2.4$\pm$1.7 &                    2.71$\pm$2.4 &                  5.12$\pm$4.1 &                3.25$\pm$2.5 &  0.51$\pm$0.4 \\
Adv black  &  0.59$\pm$0.9 &   0.2$\pm$0.2 &  \textbf{1.22$\pm$0.9} &       0.25$\pm$0.3 &                    0.26$\pm$0.3 &                  0.28$\pm$0.4 &                0.26$\pm$0.3 &  0.16$\pm$0.2 \\

Space Invaders: \\

Q-val rand     &  -0.7$\pm$0.6 &  -0.6$\pm$0.6 &  -0.8$\pm$0.6 &      \textbf{-0.39$\pm$0.4} &                   -1.12$\pm$0.9 &                 -0.81$\pm$0.7 &               -0.88$\pm$0.7 &  -1.1$\pm$0.8 \\
Adv rand &  0.76$\pm$3.5 &  0.83$\pm$3.4 &  0.79$\pm$3.7 &       0.66$\pm$2.8 &                    0.87$\pm$4.3 &                  0.76$\pm$3.7 &                0.87$\pm$3.6 &  \textbf{0.89$\pm$4.2} \\
Q-val black      &  1.01$\pm$0.2 &  0.73$\pm$0.1 &  0.74$\pm$0.2 &       0.89$\pm$0.1 &                    1.02$\pm$0.2 &                  1.08$\pm$0.2 &                \textbf{1.11$\pm$0.3} &  0.56$\pm$0.1 \\
Adv black  &  0.28$\pm$0.4 &  0.26$\pm$0.3 &  \textbf{0.59$\pm$0.4} &       0.21$\pm$0.2 &                    0.24$\pm$0.2 &                  0.25$\pm$0.2 &                0.29$\pm$0.2 &  0.13$\pm$0.2 \\

\bottomrule
\end{tabular}
\caption{The mean and SD of the insertion metric curve for $1000$ states of each game. \emph{Q-val} and \emph{Adv} measure the change of the normalized q-value and advantage respectively. \emph{Rand} and \emph{black} use random and black perturbation respectively during the insertion metric.}
\label{tb:insertion_metric}
\end{table*}

Table \ref{tb:insertion_metric} reports the \textbf{insertion metric} results for 1000 states of each game and each saliency map approach.
To get a baseline performance, we also calculated the insertion metric with random saliency maps.
For some games and sub-metrics, the mean area under the insertion curve is negative. 
This is due to the fact that some agents assign high negative q-values and advantages to the fully perturbed state.
For most games, RISE has the best results for measuring the raw q-values on random perturbation.
However, the results for measuring advantage with random perturbation are poor for all approaches. 
For Frostbite and Space Invaders the random saliency maps even performed better than all other approaches.
For the other two games, RISE has the highest values.
When using black color perturbation during the insertion metric, \occlusion{} obtained very good results for measuring the state-value, and SARFA worked best for the advantage.
However, their results for random perturbation were very poor.
From our parameter tuning, we knew that this depended on the color of perturbation used  during the saliency map generation.
Therefore, we additionally tested \occlusion{} with gray color and SARFA with noise perturbation as used by Noise Sensitivity.
With this, \occlusion{} got the highest q-value random insertion results in Pac-Man (2.98$\pm$3.5),  Frostbite (3.54$\pm$2.3), and  Space Invaders (0.07$\pm$0.7).
SARFA got the best advantage results for random insertion for Pac-Man (1.12$\pm$1.0) and Frostbite (0.73$\pm$1.2) only losing to \occlusion{} with gray color in Space Invaders (1.02$\pm$3.7 compared to 1.04$\pm$3.5).
The performance of both approaches on black perturbation fell to a level similar to the random baseline.
For the full results see Appendix \ref{ap:additonal_results}.
The exception to most observations described above is Breakout.
Here, the LIME variants performed the best across most metrics. 
SLIC segmentation in particular achieves at least the second-highest score in each metric.
Notably, this game has the highest SD values.
\section{Discussion}
\label{sec:discussion}

The results of our \textbf{sanity checks} show that most of the perturbation-based saliency map approaches tested in this paper are dependent on the learned parameters of the agent's neural network.
The results are generally comparable to the best gradient-based approaches tested by \citet{adebayo2018sanity} and the best modified propagation approaches tested in \citet{sixt2019}.
The only exceptions to this are Noise Sensitivity and LIME.

Noise Sensitivity showed little dependence on the parameters of the output layer (Fig. \ref{fig:sanity_checks}).
Since the output layer has the highest impact on the actual decision of a network, it is crucial that a faithful saliency map depends on the weights learned in this layer.
Our results empirically show that replacing the original equation of Noise Sensitivity to calculate the importance of each pixel with the equation used by Occlusion Sensitivity greatly increases the parameter dependence. 
We think that this is due to the fact that the original equation takes all actions into account and therefore measures a general increase in entropy within the activations of the output layer.  
In contrast, \occlusion{} only measures the action which is actually chosen and therefore captures a more specific change in the output layer activation.
Recently, \citet{puri2020} also criticized that the saliency maps by \citet{greydanus2018visualizing} take all actions into account. 
The results of our sanity checks provide the first computational evidence for this critique.

LIME performed well in the sanity check measurements using SSIM and Spearman correlation.
Only the Pearson correlation of the HOGs was very high between LIME saliency maps for the trained and randomized agents. 
However, the reason for this is not necessarily a low dependence on the agent's learned weights. 
More likely it is due to the fact that all LIME saliency maps for a given state work with the same superpixels. 
Since every pixel inside a superpixel has the same value there are hard edges between the superpixels. 
These edges are captured by the HOGs and result in high values of the Pearson correlation of the HOGs.

The \textbf{insertion metric} results are more nuanced.
During our parameter tuning, we tried our best to find parameters that result in saliency maps that work for both black and random perturbation and capture both the agent's action-value as well as state-value estimation. 
Despite these efforts, no saliency map approach performed well across all sub-metrics.
The best results for measuring the state-value were obtained by \occlusion{} and the best results for the action-value were obtained by SARFA.
This distinction is illustrated by the fact that no SARFA salience map, which we looked at, identified the in-game score as relevant (e.g. Fig. \ref{fig:expl_pac}).
The score is a good indicator for the value of the current state and is regularly highlighted by all other approaches we tested.

Additionally, the saliency maps' fidelity depended on the type of perturbation.
The area under the insertion curve with black perturbation was the highest when the saliency approaches used black occlusion.
To mitigate this effect some saliency approaches utilize blurring during their perturbation.
Surprisingly, this was also sensitive to the perturbation type of the insertion metric in our tests. 
Similar to gray occlusion, blurring performed best for the random perturbation insertion metric and did not do well on black perturbation.
The closest thing to a saliency map approach that fits all sub-metrics was RISE.
However, the results here were considerably worse than the results for \occlusion{} and SARFA with parameters that fit the respective sub-metric, especially when analyzing the action-value estimation.


\paragraph{Limitations}
We used four different variants of the insertion metric to get a good estimate of saliency approaches' fidelity in different situations.
Between those variants, we already found distinct differences.
This fact reinforces the findings by \citet{Tomsett20} that current fidelity metrics for saliency maps can be very sensitive to specifics of their implementation.
For value-based RL in particular, we extend the results of \citeauthor{Tomsett20} by demonstrating that there are also considerable differences between metrics that measure the action-value and metrics that measure the state-value.
However, it can not be ruled out that other fidelity metric variants might result in even more insights.
To ease future evaluations and parameter searches, a great challenge for XAI research will be the development of more general fidelity metrics for saliency maps.

Another potential limitation of our results is that recent work indicates that simply displaying saliency maps to end-users might not be suited as a final explanation \citep{huber2020LocalandGlobal,Danesh2021}.
However, saliency maps are still often used as primary components of more sophisticated explanation frameworks (e.g. \citep{Danesh2021}).
We argue that it is even more crucial to evaluate the fidelity of saliency maps in situations where their information is used as an integral component of more complex explanation mechanisms.

\section{Conclusion}
\label{sec:conclusion}

This paper compared five different perturbation-based saliency map approaches measuring their dependence on the agent's parameters and their fidelity to the agent's reasoning.

\emph{Most of the approaches tested in this work do depend on the agent's learned parameters.
Only Noise Sensitivity showed less dependence on the learned parameters of the output layer.} 
We empirically show that replacing Noise Sensitivity's original importance calculation with a calculation that only takes the chosen action into account, drastically increases parameter dependence of the output layer. 

Regarding fidelity, there is no single saliency map approach that fits every situation.
\emph{For value-based DRL agents, there are considerable differences between analyzing the agent's action-value and state-value estimation.}
While this distinction is hidden within the agent's output q-values, future practitioners should be aware of which of the two they want to analyze and choose their saliency maps accordingly. 
In our tests, SARFA worked best to  capture the action-value while \occlusion{} and RISE were more suited for the state-value.
\emph{Depending on which perturbation method the approaches use, the resulting saliency maps only analyze how sensitive the agent is with regard to specific types of perturbation.}
While this seems obvious, it was true even for perturbation methods that utilized blurring specifically to reduce their dependence on a choice of occlusion color.
In contrast to the action- and statue-value distinction, this is not an inherent property of the DRL agents but might be seen as a flaw of current perturbation-based saliency approaches.
Our results demonstrate that there is still a need to further develop perturbation-based saliency approaches.
For now, researchers have to decide which types of perturbation are meaningful and interesting for their application. 
Based on this, they can choose an appropriate perturbation method.
For example, by performing a parameter search similar to the one conducted in this work.

\bibliography{refs}
\bibliographystyle{named}

\appendix
\clearpage

\section{Saliency Map Approaches}
\label{sec:saliency_maps}
The basic saliency map generation process is the same between all four approaches compared in this work. 
Let $\pi: \mathbb{R}^{H \times W \times c} \rightarrow \mathbb{R}^{m}$ be the agent that takes a visual input state $I$ and maps it to a q-value $q(I,a)$ for each possible action.
To ease notation we use $q(I)$ to describe the q-value of the action that the agent chose in the unperturbed input $I$. 
To determine the relevance of each pixel $\lambda$ for the prediction of the agent, all five approaches feed perturbed versions of $I$ to the agent and then compare the resulting q-values with the original results.
However, the approaches widely differ in the way the input is perturbed and how the relevance per pixel is computed:

\textbf{Occlusion Sensitivity \cite{zeiler2014visualizing}:}
This approach creates perturbed states $I^\prime$ by shifting a $n \times n$ patch across the original state $I$ and occluding this patch by setting all the pixels within to a certain color (e.g., black).
The importance $S(\lambda)$ of each pixel $\lambda$ inside the patch is then computed based on the agents' confidence after the perturbation
\begin{equation}
\label{eq:occlusion}
    S(\lambda) = 1 - q(I').
\end{equation}
Since the original source does not go into details about the algorithm, we use the \emph{tf-explain} implementation as reference\footnote{Available under: \url{https://github.com/sicara/tf-explain}}.
As long as the saliency maps are normalized this is equivalent to $q(I) - q(I^{\prime})$ since all values in the saliency map are shifted by the same constant $q(I)-1$. 

\textbf{Noise Sensitivity \citep{greydanus2018visualizing}:}
Instead of completely occluding patches of the state, this approach adds noise to the state $I$ by applying a Gaussian blur to a circle with radius $r$  around a pixel $\lambda$.
The modified state $I^{\prime}(\lambda)$ is then used to compute the importance of the covered circle by comparing the agent's logit units $\pi(I)$:
\begin{equation}
    \label{eq:noise}
    S(\lambda) = \frac{1}{2}||\pi(I) - \pi(I^{\prime}(\lambda))||^2
\end{equation}
This is done for every $r$th pixel, resulting in a temporary saliency map smaller than the input. 
For the final saliency map, the result is up-sampled using bilinear interpolation.

\textbf{RISE \citep{petsiuk2018rise}:} 
This approach uses a set of $N$ randomly generated masks $\{M_1,..., M_N\}$ for perturbation.
To this end, temporary $n \times n$ masks are created by setting each element to $1$ with a probability $p$ and $0$ otherwise. 
These temporary masks are upsampled to the size of the input state using bilinear interpolation.
The states are perturbed by element-wise multiplication with those masks $I \odot M_i$.
The relevance of each pixel $\lambda$ is given by
\begin{equation}
    S(\lambda) = \frac{1}{p \cdot N} \sum \limits_{i=1}^{N} q(I \odot M_i) \cdot M_i(\lambda),
\end{equation}
where $M_i(\lambda)$ denotes the value of the pixel $\lambda$ in the $i$th mask.

\textbf{LIME \citep{ribeiro2016should}:} 
The original state is divided into superpixels using segmentation algorithms.
Perturbed variations of the state are generated by ``deleting'' different combinations of superpixels (i.e., setting all pixels of the superpixels to $0$).
The combination of occluded states and the corresponding predictions by the agent are then used to train a locally weighted interpretable model for $N$ steps.
Analyzing the weights of this local model provides a relevance value for each superpixel.

\textbf{SARFA \citep{puri2020}:}
This approach does not use a specific perturbation method. 
\citeauthor{puri2020} test noise perturbation for Atari games and occlusion for other domains.
Given a perturbed sate $I'$, SARFA measures the information specific to the chosen action $a'$ by calculating $\Delta p = softmax(\pi(I))_{a'} - softmax(\pi(I'))_{a'}$.
To only measure relevant information, they additionally calculate $K=KL(softmax_{a'}(\pi(I')), softmax_{a\prime}(\pi(I)))$, where  $KL$ is the Kullback–Leibler divergence and $softmax_{a'}$ is the softmax over all outputs \emph{except} the chosen action $a'$.
The final score for the perturbed state $I'$ is then given by:
\begin{equation}
    S(\lambda) =
    \frac{2K\Delta p}{K + \Delta p}
\end{equation}

\begin{figure*}[ht]
\centering
\newcommand{\mysize}{0.13} 
\newcommand{\myskip}{2\baselineskip} 
 \begin{minipage}[t]{\mysize\linewidth}
    \parbox[t][\myskip]{\linewidth}{\centering
        Input State}
        \includegraphics[width=\linewidth]{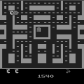}
    \end{minipage}
    \begin{minipage}[t]{\mysize\linewidth}
    \parbox[t][\myskip]{\linewidth}{\centering
        Black Occlusion}
         \includegraphics[width=\linewidth]{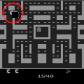}
    \end{minipage}
    \begin{minipage}[t]{\mysize\linewidth}
    \parbox[t][\myskip]{\linewidth}{\centering
        Grey Occlusion}
        \includegraphics[width=\linewidth]{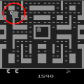}
    \end{minipage}
    \begin{minipage}[t]{\mysize\linewidth}
    \parbox[t][\myskip]{\linewidth}{\centering
       Noise Sensitivity}
        \includegraphics[width=\linewidth]{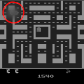}
    \end{minipage}
    \begin{minipage}[t]{\mysize\linewidth}
    \parbox[t][\myskip]{\linewidth}{\centering
        RISE}
        \includegraphics[width=\linewidth]{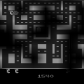}
    \end{minipage}
    \begin{minipage}[t]{\mysize\linewidth}
    \parbox[t][\myskip]{\linewidth}{\centering
        LIME Segmentation}
        \includegraphics[width=\linewidth]{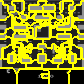}
    \end{minipage}
    \begin{minipage}[t]{\mysize\linewidth}
    \parbox[t][\myskip]{\linewidth}{\centering
        LIME Perturbed}
         \includegraphics[width=\linewidth]{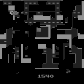}
    \end{minipage}
    
\caption{An example of the different types of perturbation used by the saliency map approaches in our work. The parameters are chosen in such a way that the idea of the perturbation can be easily identified. For Occlusion and Noise, the disturbed area is marked with a red circle.}
\label{fig:perturbation_examples}
\end{figure*}

Fig. \ref{fig:perturbation_examples} shows an example for the different perturbation methods used by the saliency map approaches in this work.

\section{Tuning The Metrics}

\subsection{Calibration for Similarity Metrics}
\label{ap:similarity_metric_test}
The sanity checks use three similarity metrics: Spearman rank correlation, Structural Similarity (SSIM), and Pearson correlation of the Histogram of Oriented Gradients (HOGs). We need to calibrate these metrics such that high similarity values actually indicate similar saliency maps.
Analogous to \citet{adebayo2018sanity}, we do this by calculating the similarity of $100$ pairs of randomly generated saliency maps (Uniform and Gaussian).
Since randomly sampled saliency maps should be very different on average, the mean of these similarities should be low.
Using an SSIM window size of 7 and a HOG function with $(3, 3)$ pixels per cell, two randomly sampled saliency maps with uniform distribution had mean similarity values $(0.0087,0.0136, 0.0096)$ and two random saliency maps with Gaussian distribution had mean similarity $(0.0093, 0.0374, 0.0087)$.

\subsection{Normalizing The Insertion Metric Results}
\label{ap:normalizing_insertion_metric}
To be more robust against outlier states, we tested two different ways to normalize the q-values and advantage values during the insertion metric.
The first normalization method we tested was inspired by \citet{sixt2019} and forces each insertion curve to start at $0$ and finish at $1$.
This is achieved by applying $f(x) = \frac{x - b}{t-b}$ to each insertion step result, where $b$ is the output of the fully perturbed state and $t$ is the output of the original state.
As the second method, we only divided each insertion step by the output of the original state $t$.
In this way, all insertion curves finish at the value of $1$.
Table \ref{tb:appendix_normalization} shows the range of standard deviations (SD) of the area under the insertion curve of 28 variants of Occlusion Sensitivity saliency maps on 1000 Pac-Man states for each normalization method. 
Interestingly, the full normalization to curves between $0$ and $1$ resulted in the highest SD.
We think that this comes from the fact that our agents sometimes assign higher values to the fully perturbed state than to the original state.
In these cases, $t-b$ is negative, and applying $f(x)$ actually inverts the insertion curve.

\begin{table}
\centering
\begin{tabular}{lrr}
\toprule
Normalization Function &	Minimum SD & Maximum SD \\
\midrule
Measuring Q-Values \\
No Normalization & 5.16 & 10.17 \\
$f(x)=\frac{x}{t}$ & 1.14 & 2.06 \\
$f(x)=\frac{x - b}{t-b}$ & 10.33 & 48.56 \\
Measuring Advantage \\
No Normalization & 0.84 & 1.42 \\
$f(x)=\frac{x}{t}$ & 1.99 & 3.78 \\
$f(x)=\frac{x - b}{t-b}$ & 9.45 & 165.20 \\

\bottomrule
\end{tabular}
\caption{The minimum and maximum SD when evaluating 28 different parameter combinations of Occlusion Sensitivity saliency maps with an insertion metric using different normalization functions.}
\label{tb:appendix_normalization}
\end{table}

\section{Parameter Tuning Details}

\subsection{State selection}
\label{ap:parameter_tuning_state_selection}
To tune the parameters of our perturbation-based saliency map approaches, we needed to find a subset of $10$ states that represent the 1000 states used in our final evaluation.
In addition to 10 randomly generated subsets, we used the HIGHLIGHTS-DIV algorithm to generate subsets.
HIGHLIGHTS-DIV assigns an importance value to each state that measures how important that state was for the agent's strategy. 
Then the states with the highest importance values which are additionally less similar than a predefined diversity threshold are chosen for the subset. 
For the diversity threshold, we tested the 10,20,25,28,30,32,33,35, and 40 percentile of the distribution of the similarity values of the full 1000 states stream.
To get even more diverse sets of states, we additionally tested two novel variations of HIGHLIGHTS-DIV: one variant where we used 5 of the most important and 5 of the least important states and one variant where we sorted all 1000 states by importance and chose every 100th state to obtain states of all importance levels.

The highest correlation to the ranking obtained by the full stream was achieved by the 30 percentile HIGHLIGHTS-DIV variant. 
For the action-value, the Spearman's rank correlation was $0.96$ and the Kendall rank correlation was $0.85$. 
For the state-value, the Spearman's rank correlation was $0.95$ and the Kendall rank correlation was $0.81$. 
The correlations for the other subsets can be seen in our repository. \footnote{
\url{https://github.com/belimmer/PerturbationSaliencyEvaluation}}

\subsection{Saliency map parameters.}
\label{ap:parameter_tuning_used_parameters}
For all perturbation-based saliency maps, we tested several parameter combinations as described in Section \ref{sec:parameter_tuning}. As final parameters, we chose the parameters that obtained the best combined insertion metric AUC and needed a maximum of 3 seconds to compute each saliency map. 
The full results of our tests can be viewed in our repository. \footnote{
\url{https://github.com/belimmer/PerturbationSaliencyEvaluation}}

For Occlusion Sensitivity, we tested patches of size $1$ to $10$, black and gray occlusion color, and whether applying a softmax layer to the output q-values before creating the saliency map improves results.
The top five results are shown in Table \ref{tab:appendix_occl}. 

For Noise Sensitivity, we tested circles with a radius of $1$ to $10$. 
The top five parameters are shown in Table \ref{tab:appendix_noise}.

SARFA was not introduced with a specific perturbation method.
Analogous to \citeauthor{puri2020}, we test blurred circles of radius $1$ to $10$ as used in Noise Sensitivity. 
Additionally, we also use circles that are occluded with black color.
The top five results are shown in Table \ref{tab:appendix_sarfa}.

For RISE we tested $500$, $1000$,...,$3000$ masks of size $4$ to $24$.
The probability $p$ with which each pixel is occluded varied between $0.1$ and $0.9$ in steps of $0.1$.
Analogous to Occlusion Sensitivity, we also investigated whether it makes sense to add a softmax layer after the output during the saliency map creation.
The top five results are shown in Table \ref{tab:appendix_rise}.

For LIME we tested the three most common Segmentation techniques \emph{SLIC}, \emph{Quickshift} and \emph{Felzenszwalb} and varied the number of samples on which the local interpretable model is trained. 
For the number of learning steps we took the default number of samples ($1000$) and increased it in steps of $500$ up to $3000$.
To determine which parameter ranges we should explore for each segmentation algorithm, we performed preliminary tests where we visually checked which parameters resulted in different segmentation. 
For Felzenszwalb segmentation we used a scale factor of $1$,$21$,...,$101$, a minimum component size from $1$ to $8$ and Gaussian smoothing kernels with width $\sigma$ of $0$,$0.25$,...,$1$.
The top results are shown in Table \ref{tab:appendix_felzenszwalb}.
For SLIC we tested $40$,$60$ to $240$ segments, a compactness factor of $0.001$,$0.01$,...,$10$ and Gaussian smoothing kernels with width $\sigma$ of $0$,$0.25$,...,$1$.
The top five parameter combinations can be seen in Table \ref{tab:appendix_slic}.
Finally, we tested Quickshift with a color ratio of $0.0$,$0.33$,$0.66$ and $0.99$, a kernel size from $1$ to $6$ and a max distance of $kernel size * i$, where $i$ goes from $1$ to $4$.
The top results are shown in Table \ref{tab:appendix_quickshift}.

\begin{table}[ht]
        \centering
        \begin{tabular}{ccccc}
        \toprule
        AUC & Patch Size & Color & Softmax & Time
        \\  
        \midrule
        6.76 & 1 & Black & No & 10.94 \\  
        3.44 & 1 & Gray & No & 11.09 \\  
        3.42 & 1 & Black & Yes & 11.51 \\  
        3.03 & 1 & Gray & Yes & 11.50 \\  
        \textbf{2.33} & \textbf{2} & \textbf{0.0} & \textbf{No} & \textbf{2.80} \\  
        \bottomrule
        \end{tabular}
        \caption{Best parameters for Occlusion Sensitivity. The final parameters are marked in bold.}
        \label{tab:appendix_occl}
\end{table}

\begin{table}[ht]
        \centering
        \begin{tabular}{ccc}
        \toprule
        AUC & Radius  & Time
        \\
        \midrule
        3.08 & 2 & 5.79 \\   
        2.21 & 1 & 22.84 \\   
        \textbf{0.94} & \textbf{3} & \textbf{2.62} \\   
        0.68 & 9 & 0.38 \\   
        0.48 & 10 & 0.31 \\ 
        \bottomrule
        \end{tabular}
        \caption{Best parameters for Noise Sensitivity. The final parameters are marked in bold.}
        \label{tab:appendix_noise}
\end{table}

\begin{table}[ht]
        \centering
        \begin{tabular}{cccc}
        \toprule
        AUC & Radius & Perturbation & Time
        \\
        \midrule
        7.03 & 1 & Black & 12.05 \\   
        \textbf{1.46} & \textbf{2} & \textbf{Black} & \textbf{3.00} \\   
        1.09 & 1 & Blur & 23.70 \\   
        0.57 & 8 & Blur & 0.46 \\   
        0.55 & 2 & Blur & 6.12 \\ 
        \bottomrule
        \end{tabular}
        \caption{Best parameters for SARFA. The final parameters are marked in bold.}
        \label{tab:appendix_sarfa}
\end{table}

\begin{table}[ht]
    \newcommand{\mysize}{0.11\linewidth}
    \centering
    \begin{tabular}{p{\mysize}p{\mysize}p{\mysize}p{\mysize}p{\mysize}p{\mysize}}
    \toprule
    AUC & $p$ & Mask Size & Masks & Softmax  & Time
    \\   
    \midrule
3.21 & 0.8 & 11 & 3000 & Yes & 5.09 \\   
3.04 & 0.7 & 13 & 3000 & No & 4.76 \\    
2.99 & 0.9 & 24 & 2500 & Yes & 3.98 \\   
2.94 & 0.8 & 4 & 3000 & No & 4.66 \\  
\vdots & \multicolumn{5}{c}{\parbox{0.55\linewidth}{Skipping 9 parameters that took more then 3 seconds.}} \\
\textbf{2.66} & \textbf{0.5} & \textbf{8} & \textbf{1000} & \textbf{No} & \textbf{1.54} \\   
    \bottomrule
    \end{tabular}
    \caption{Best parameters for RISE. The final parameters are marked in bold. }
    \label{tab:appendix_rise}
\end{table}

\begin{table}[ht]
    \newcommand{\mysize}{0.11\linewidth}
    \centering
    \begin{tabular}{p{\mysize}p{\mysize}p{\mysize}p{\mysize}p{\mysize}p{\mysize}}
    \toprule
    AUC & Scale & Sigma & \hspace{0pt} Minimum Size & Num Samples & Time \\ 
    \midrule
    4.35 & 21 & 0.5 & 0 & 3000 & 10.73 \\
    3.58 & 21 & 0.75 & 2 & 3000 & 7.38 \\
    3.53 & 1 & 1.0 & 0 & 2000 & 22,03 \\
    3.29 & 21 & 0.5 & 0 & 2500 & 8.95 \\
\vdots & \multicolumn{5}{c}{\parbox{0.55\linewidth}{Skipping 14 parameters that took more then 3 seconds.}} \\
    \textbf{2.55} & \textbf{21} & \textbf{0.5} & \textbf{4} & \textbf{1000} & \textbf{1.71} \\
    \bottomrule
    \end{tabular}
 \caption{Best parameters for LIME with Felzenszwalb segmentation. The final parameters are marked in bold.
 }
    \label{tab:appendix_felzenszwalb}
\end{table}

\begin{table}[ht]
    \newcommand{\mysize}{0.11\linewidth}
    \centering
    \begin{tabular}{p{\mysize}p{\mysize}p{\mysize}p{\mysize}p{\mysize}p{\mysize}}
    \toprule
    AUC & Number of Segments & \hspace{0pt} Compactness & Sigma & Num Samples & Time
    \\   
    \midrule
    3.99 & 200 & 10.0 & 1.0 & 3000 & 3.13 \\
\textbf{3.86} & \textbf{200} & \textbf{10.0} & \textbf{0.25} & \textbf{2000} & \textbf{2.08} \\
3.48 & 200 & 10.0 & 0.0 & 3000 & 3.11 \\
3.46 & 200 & 0.001 & 0.25 & 3000 &	2.36 \\
3.44 & 200 & 10.0 & 0.5 & 1000 & 1.06 \\
    \bottomrule
    \end{tabular}
    \caption{Best parameters for LIME with SLIC segmentation. The final parameters are marked in bold.}
    \label{tab:appendix_slic}
\end{table}

\begin{table}[ht]
    \newcommand{\mysize}{0.11\linewidth}
    \centering
    \begin{tabular}{p{\mysize}p{\mysize}p{\mysize}p{\mysize}p{\mysize}p{\mysize}}
     \toprule
    AUC & Kernel Size & Max Distance & Ratio & Num Samples & Time
    \\   
    \midrule
    6.24 & 1 & 1 & 0.0 & 3000 &	11.38 \\
4.97 & 1 & 1 & 0.0 & 2500 & 9.57 \\
4.80 & 1 & 2 & 0.0 & 2500 & 4.46 \\
4.50 & 1 & 2 & 0.0 & 3000 &	5.39 \\
\textbf{4.23} & \textbf{1} & \textbf{2} & \textbf{0.0} & \textbf{1500} & \textbf{2.75} \\
    \bottomrule
    \end{tabular}
    \caption{Best parameters for LIME with Quickshift segmentation. The final parameters are marked in bold.}
    \label{tab:appendix_quickshift}
\end{table}

\section{Run-time Analysis}
\label{ap:runtime}

The run-time of an algorithm can be an important aspect when choosing between different approaches. 
We computed the mean time it took each algorithm to create a single saliency map using the $timeit$ python library.
To get a feeling of how this is affected by different parameters of the saliency approaches, we measured the time during our parameter tuning process where each parameter combination was used on $10$ different states.

The fastest approach was Occlusion Sensitivity which uses simple color occlusions followed by the more complex blur perturbation of SARFA and Noise Sensitivity. 
However, this was strongly dependent on the size of the perturbation patches and circles respectively. 
Using a patch size or radius of $1$, these approaches were among the slowest with a mean run-time of around $22s$ for the blur perturbation and approximately $11s$ for the black occlusion variant.
However, increasing the patch size and radius to $2$ already drastically reduced the run-time as can be seen in the tables of Section \ref{ap:parameter_tuning_used_parameters}.

For RISE, the run-time mainly depends on the number of masks. 
With $3000$ masks the run-time was always close to $5s$ per saliency map.
However, compared to the aforementioned saliency approaches, this did only decrease slowly when decreasing the number of masks.
Thus, the average and the fastest run-time were much slower for RISE than for SARFA, and Occlusion and Noise Sensitivity.

The slowest approach we tested was LIME.
However, this was strongly influenced by the number of segments that the segmentation functions generated and the number of learning steps for the locally interpretable classifier.
For SLIC, which creates relatively big segments, LIME was quite fast with a maximum run-time of $3.87s$ with the slowest parameters.
In contrast, the run-time for Felzenswalb easily exploded and reached a maximum of $33.64s$ per saliency map.
Quickshift was in the middle of those two approaches with a maximum run-time of $12.50s$ which did not decrease as quickly as the run-time of Occlusion and Noise Sensitivity, and SARFA.  

The exact run-times for our top parameters can be seen in Appendix \ref{ap:parameter_tuning_used_parameters} and the values for all the parameters we tested can be found in our repository \footnote{
\url{https://github.com/belimmer/PerturbationSaliencyEvaluation}}

\section{Additional Results}
\label{ap:additonal_results}

\begin{figure*}[ht]
\centering
\newcommand{\mysize}{0.115} 
\newcommand{\myskip}{2\baselineskip} 
 \begin{minipage}[t]{\mysize\linewidth}
    \parbox[t][\myskip]{\linewidth}{\centering
        Input State}
        \includegraphics[width=\linewidth]{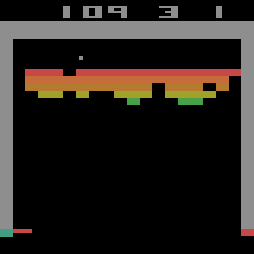}
        \includegraphics[width=\linewidth]{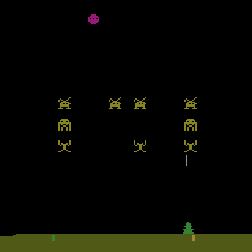}
        \includegraphics[width=\linewidth]{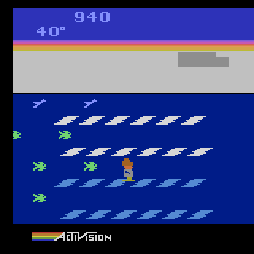}
    \end{minipage}
    \begin{minipage}[t]{\mysize\linewidth}
    \parbox[t][\myskip]{\linewidth}{\centering
        Occlusion Sensitivity}
         \includegraphics[width=\linewidth]{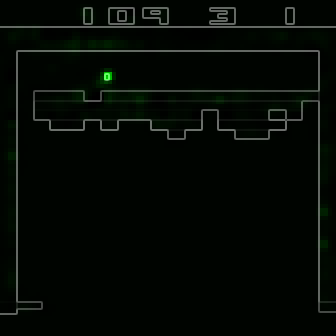}
         \includegraphics[width=\linewidth]{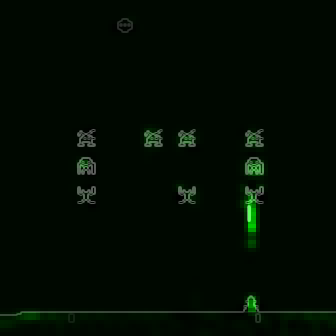}
         \includegraphics[width=\linewidth]{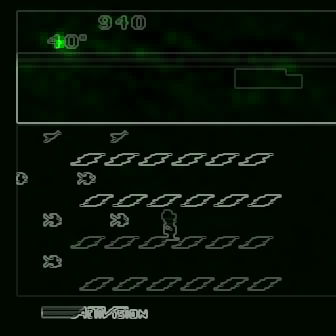}
    \end{minipage}
    \begin{minipage}[t]{\mysize\linewidth}
    \parbox[t][\myskip]{\linewidth}{\centering
       \emph{NS}  Original}
        \includegraphics[width=\linewidth]{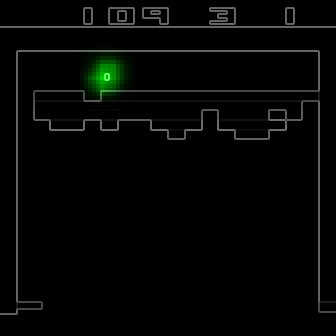}
         \includegraphics[width=\linewidth]{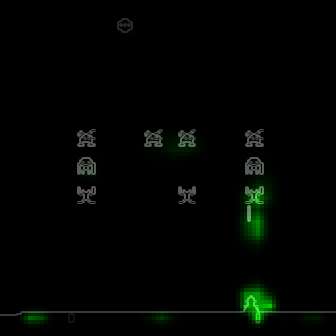}
         \includegraphics[width=\linewidth]{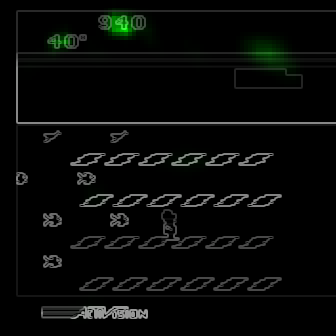}
    \end{minipage}
    \begin{minipage}[t]{\mysize\linewidth}
    \parbox[t][\myskip]{\linewidth}{\centering
        SARFA}
        \includegraphics[width=\linewidth]{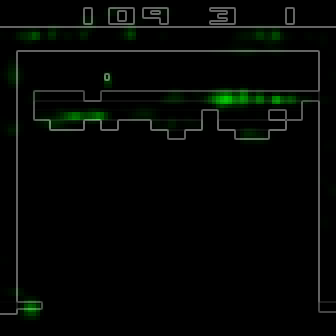}
         \includegraphics[width=\linewidth]{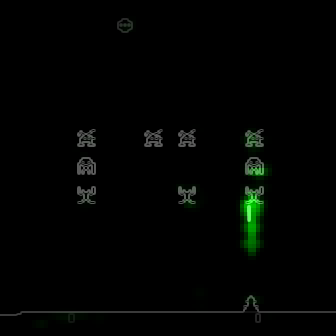}
         \includegraphics[width=\linewidth]{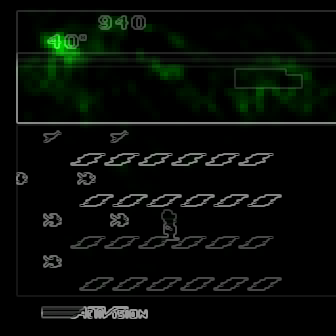}
    \end{minipage}
    \begin{minipage}[t]{\mysize\linewidth}
    \parbox[t][\myskip]{\linewidth}{\centering
        RISE}
        \includegraphics[width=\linewidth]{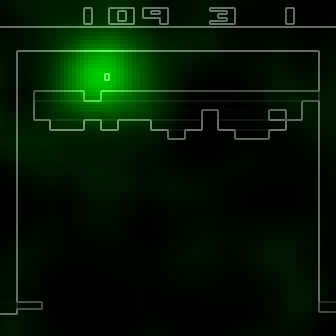}
         \includegraphics[width=\linewidth]{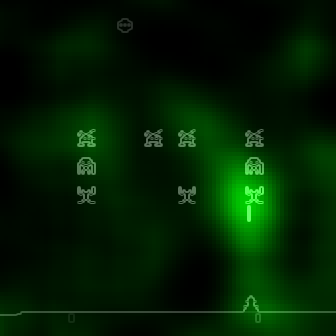}
         \includegraphics[width=\linewidth]{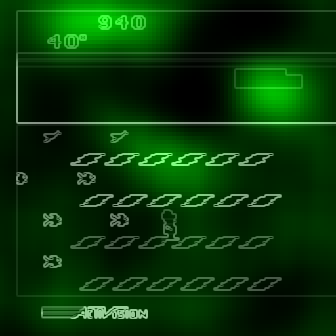}
    \end{minipage}
    \begin{minipage}[t]{\mysize\linewidth}
    \parbox[t][\myskip]{\linewidth}{\centering
        LIME Felzens.}
         \includegraphics[width=\linewidth]{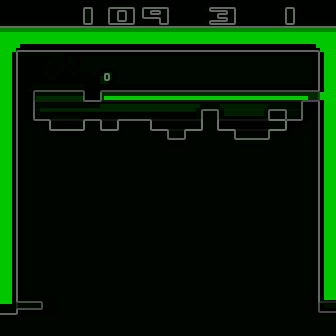}
         \includegraphics[width=\linewidth]{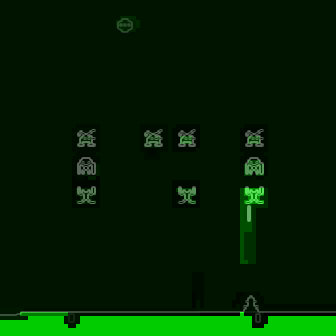}
         \includegraphics[width=\linewidth]{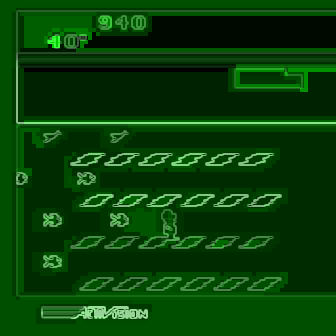}
    \end{minipage}
    \begin{minipage}[t]{\mysize\linewidth}
    \parbox[t][\myskip]{\linewidth}{\centering
        LIME Quickshift}
        \includegraphics[width=\linewidth]{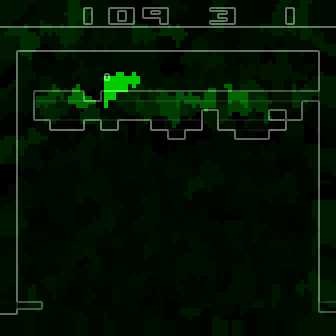}
         \includegraphics[width=\linewidth]{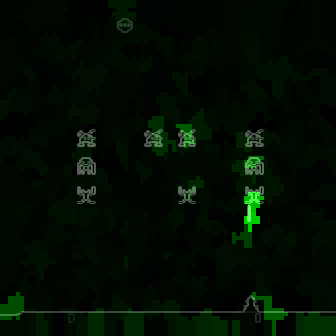}
         \includegraphics[width=\linewidth]{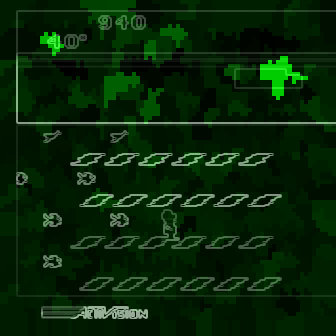}
    \end{minipage}
    \begin{minipage}[t]{\mysize\linewidth}
    \parbox[t][\myskip]{\linewidth}{\centering
        LIME SLIC}
        \includegraphics[width=\linewidth]{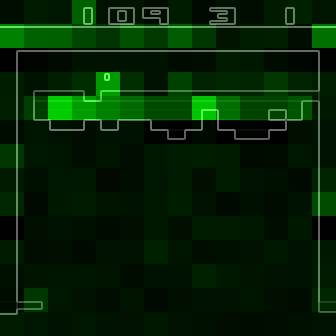}
         \includegraphics[width=\linewidth]{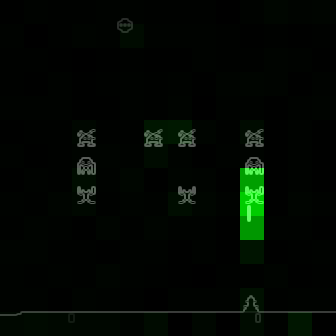}
         \includegraphics[width=\linewidth]{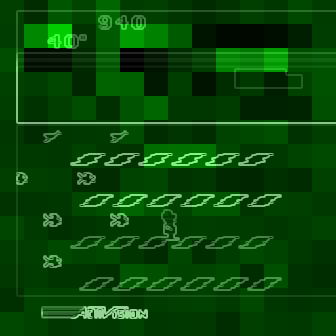}
    \end{minipage}
    
\caption{Example saliency maps for the remaining games we tested. From top to bottom: Breakout, Space Invaders and Frostbite.
\emph{NS} is Noise Sensitivity.}
\label{fig:additional_examples}
\end{figure*}

\begin{figure*}
\centering
\newcommand{\mysize}{0.11}
\newcommand{\myimagesize}{1.3cm}
\newcommand{\myskip}{2\baselineskip} 
    \begin{minipage}[t]{\mysize\linewidth}
    \vspace{\baselineskip}
        \centering
        \newcommand{\texthoehe}{\myimagesize}
         \parbox[c][\myimagesize]{\linewidth}{\centering
        Original Saliency Map}
         \parbox[c][\myimagesize]{\linewidth}{\centering
        Randomized up to FC2}
         \parbox[c][\myimagesize]{\linewidth}{\centering
        Randomized up to FC1}
         \parbox[c][\myimagesize]{\linewidth}{\centering
        Randomized up to Conv3}
         \parbox[c][\myimagesize]{\linewidth}{\centering
        Randomized up to Conv2}
         \parbox[c][\myimagesize]{\linewidth}{\centering
        Randomized up to Conv1}
    \end{minipage}
    \begin{minipage}[t]{\mysize\linewidth}
    \parbox[t][\myskip]{\linewidth}{\centering
        Occlusion}
        \centering
        \includegraphics[width=\myimagesize]{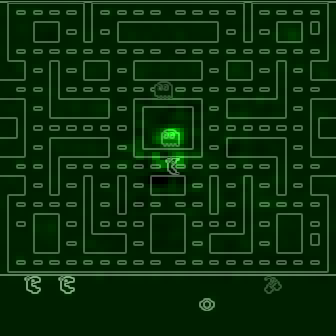}
        \includegraphics[width=\myimagesize]{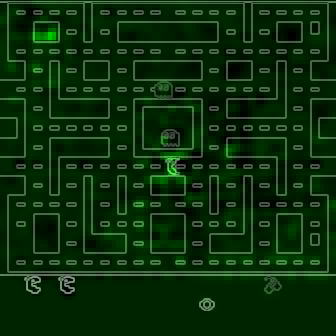}
        \includegraphics[width=\myimagesize]{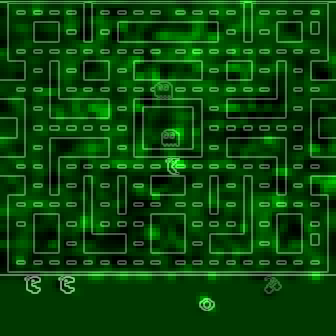}
        \includegraphics[width=\myimagesize]{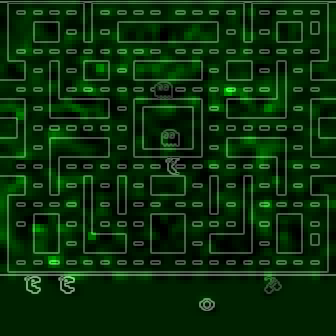}
        \includegraphics[width=\myimagesize]{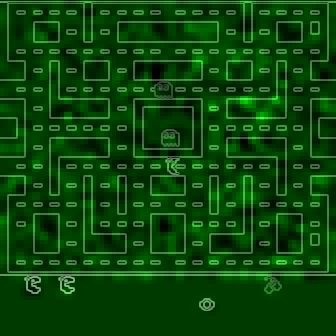}
        \includegraphics[width=\myimagesize]{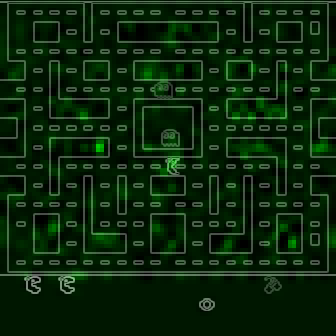}
    \end{minipage}
    \begin{minipage}[t]{\mysize\linewidth}
    \parbox[t][\myskip]{\linewidth}{\centering
        \emph{NS} Original}
        \centering
       \includegraphics[width=\myimagesize]{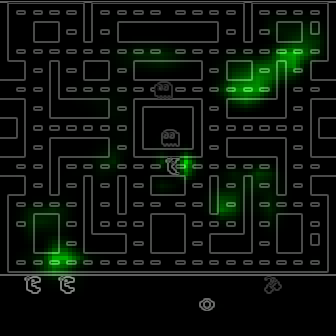}
        \includegraphics[width=\myimagesize]{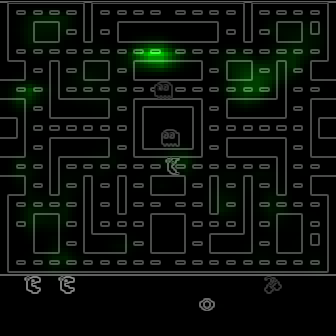}
        \includegraphics[width=\myimagesize]{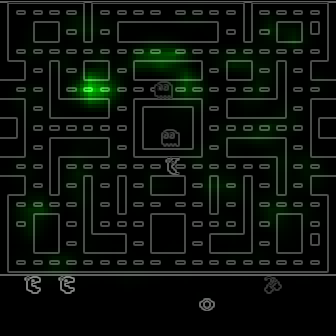}
        \includegraphics[width=\myimagesize]{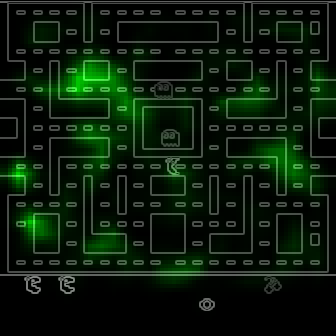}
        \includegraphics[width=\myimagesize]{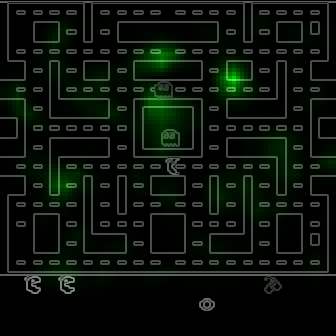}
        \includegraphics[width=\myimagesize]{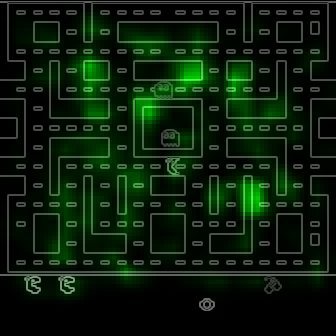}
    \end{minipage}
    \begin{minipage}[t]{\mysize\linewidth}
    \parbox[t][\myskip]{\linewidth}{\centering
        SARFA}
        \centering
      \includegraphics[width=\myimagesize]{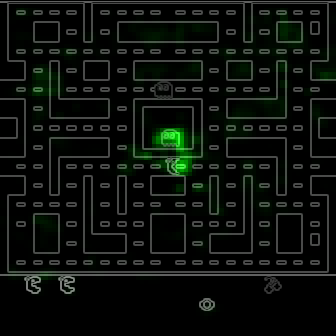}
        \includegraphics[width=\myimagesize]{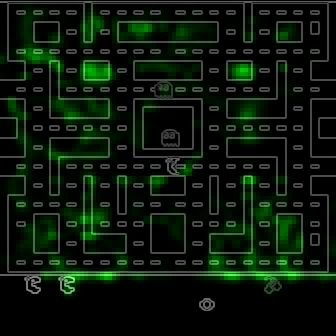}
        \includegraphics[width=\myimagesize]{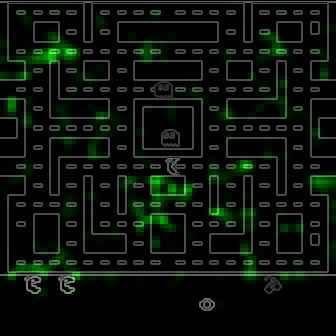}
        \includegraphics[width=\myimagesize]{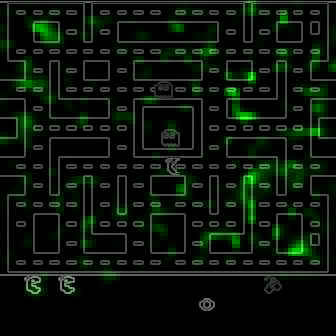}
        \includegraphics[width=\myimagesize]{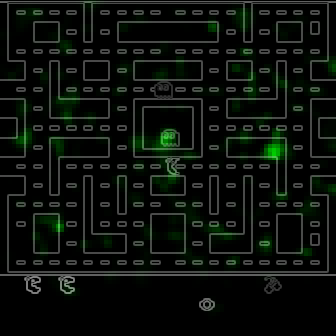}
        \includegraphics[width=\myimagesize]{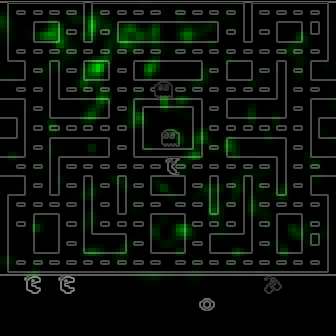}
    \end{minipage}
    \begin{minipage}[t]{\mysize\linewidth}
    \parbox[t][\myskip]{\linewidth}{\centering
        RISE}
        \centering
         \includegraphics[width=\myimagesize]{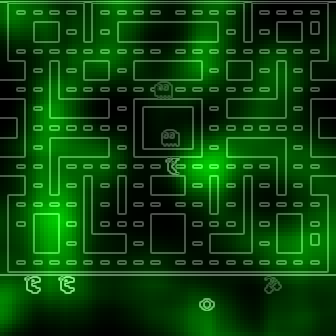}
        \includegraphics[width=\myimagesize]{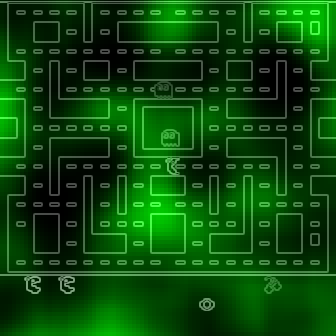}
        \includegraphics[width=\myimagesize]{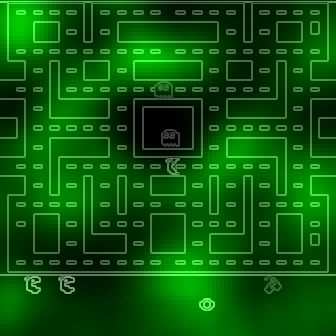}
        \includegraphics[width=\myimagesize]{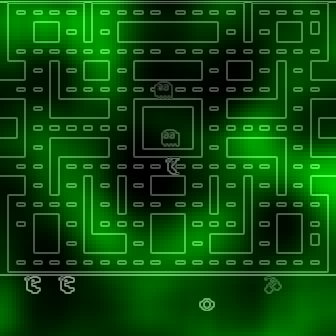}
        \includegraphics[width=\myimagesize]{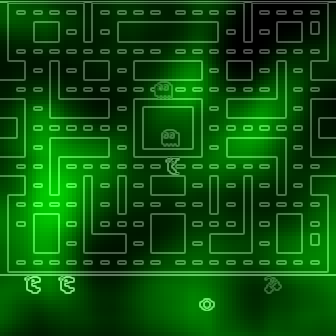}
        \includegraphics[width=\myimagesize]{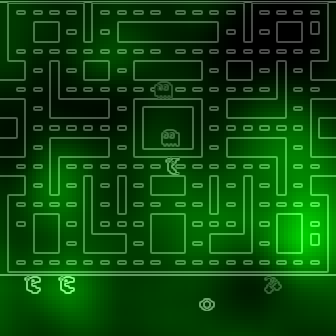}
    \end{minipage}
    \hspace{-7pt}
     \begin{minipage}[t]{0.15\linewidth}
    \parbox[t][\myskip]{\linewidth}{\centering LIME Felzenszwalb}
        \centering
         \includegraphics[width=\myimagesize]{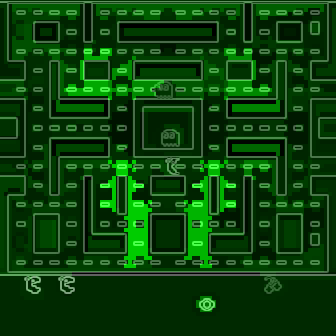}
        \includegraphics[width=\myimagesize]{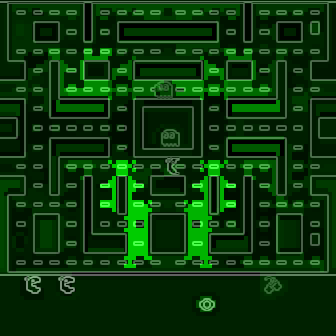}
        \includegraphics[width=\myimagesize]{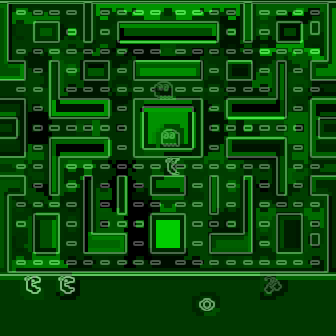}
        \includegraphics[width=\myimagesize]{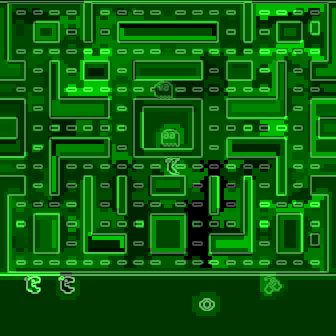}
        \includegraphics[width=\myimagesize]{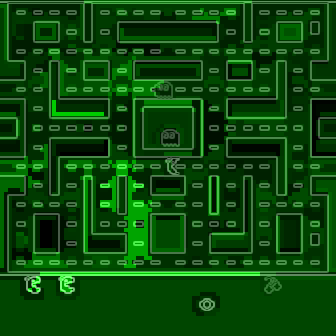}
        \includegraphics[width=\myimagesize]{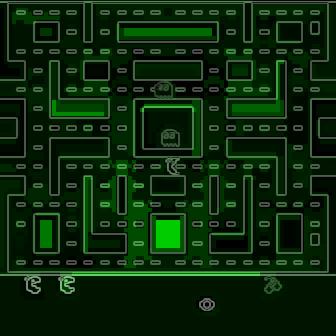}
    \end{minipage}
    \begin{minipage}[t]{\mysize\linewidth}
    \parbox[t][\myskip]{\linewidth}{\centering
        LIME Quickshift}
        \centering
        \includegraphics[width=\myimagesize]{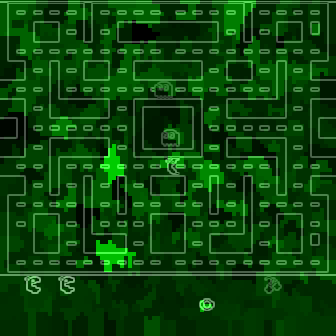}
        \includegraphics[width=\myimagesize]{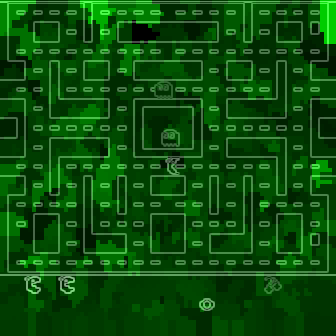}
        \includegraphics[width=\myimagesize]{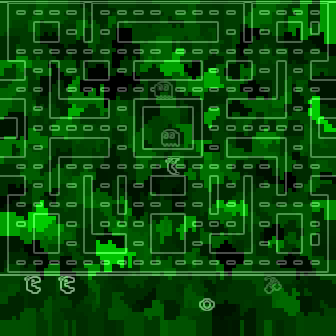}
        \includegraphics[width=\myimagesize]{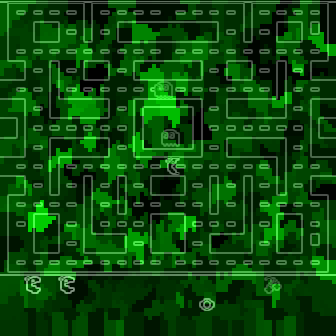}
        \includegraphics[width=\myimagesize]{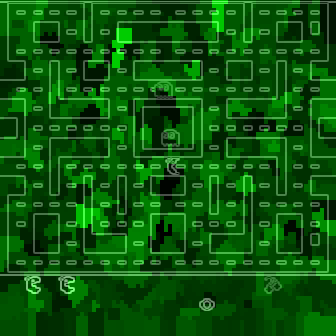}
        \includegraphics[width=\myimagesize]{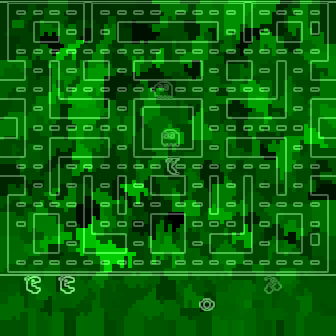}
    \end{minipage}
    \begin{minipage}[t]{\mysize\linewidth}
    \parbox[t][\myskip]{\linewidth}{\centering
        LIME SLIC}
        \centering
         \includegraphics[width=\myimagesize]{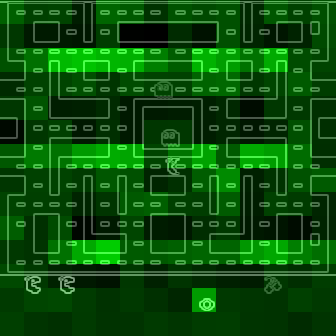}
        \includegraphics[width=\myimagesize]{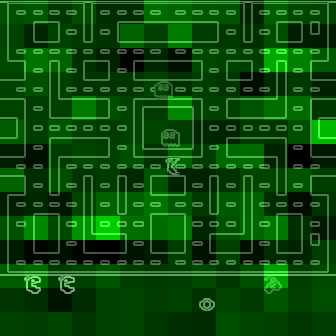}
        \includegraphics[width=\myimagesize]{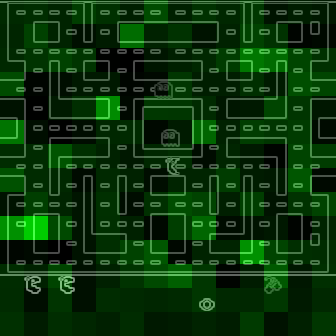}
        \includegraphics[width=\myimagesize]{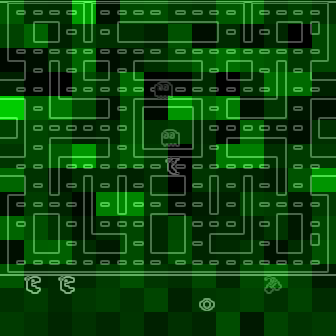}
        \includegraphics[width=\myimagesize]{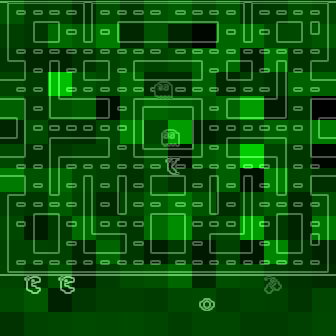}
        \includegraphics[width=\myimagesize]{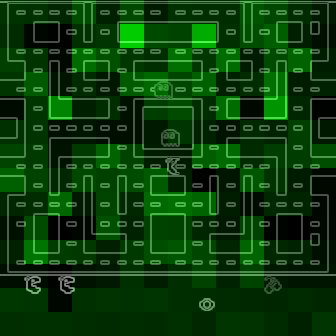}
    \end{minipage}
    
\caption{Example saliency maps for the parameter randomization sanity check.
From top to bottom each row after the first is generated for agents with cascadingly randomized layers starting with the output layer.}
\label{fig:sanity_check_example}
\end{figure*}

In this section, we show some additional results that did not fit in the main paper.
Fig. \ref{fig:additional_examples} shows example saliency maps for HIGHLIGHT-DIV states of the remaining three games apart from Pac-Man.
Fig. \ref{fig:sanity_check_example} shows an example of the saliency maps generated during a sanity check.

To properly investigate observations made during the parameter tuning, we additionally calculated the insertion metric for Occlusion Sensitivity with gray occlusion and SARFA with blur. 
All other parameters were the same as the ones used in section \ref{sec:results}.
The results are shown in Table~\ref{tb:appendix_insertion_metric}.

\begin{table}
\centering
\begin{tabular}{lrr}
\toprule
Metric &	Occlusion gray	&	SARFA blur \\
\midrule
Pac-Man: \\
Q-val rand     &    \textbf{2.98$\pm$3.5} &   1.0$\pm$2.2  \\
Adv rand &  0.44$\pm$1.8 &  \textbf{1.12$\pm$1.0} \\
Q-val black      &   0.32$\pm$0.2 & 0.62$\pm$1.3 \\
Adv black  &   -0.13$\pm$0.3 & 0.23$\pm$0.4 \\

Breakout: \\

Q-val rand & -0.83$\pm$2.6 & -0.8$\pm$3.4\\
Adv rand &  0.4$\pm$5.3 & -0.4$\pm$5.8 \\
Q-val black &   1.99$\pm$2.5 & 3.0$\pm$3.9  \\
Adv black  &   0.11$\pm$0.5 & 0.21$\pm$0.7 \\

Frostbite: \\

Q-val rand     &  \textbf{3.54$\pm$2.3} &  1.13$\pm$1.2  \\
Adv rand &   0.66$\pm$1.1 & \textbf{0.73$\pm$1.2} \\
Q-val black     &   0.5$\pm$0.5 &  0.58$\pm$0.3  \\
Adv black  &  0.16$\pm$0.2 & 0.3$\pm$0.3 \\

Space Invaders: \\

Q-val rand     &  \textbf{0.07$\pm$0.7} & -0.75$\pm$0.7 \\
Adv rand &  \textbf{1.04$\pm$3.5} &  1.02$\pm$3.7 \\
Q-val black      &  0.48$\pm$0.2 &  0.66$\pm$0.2\\
Adv black  &  0.12$\pm$0.3 &  0.44$\pm$0.4 \\

\bottomrule
\end{tabular}
\caption{The mean and SD of the insertion metric curve for our additional experiments with different perturbations for Occlusion Sensitivity and SARFA. \emph{Q-val} and \emph{Adv} measure the change of the normalized q-value and advantage respectively. \emph{Rand} and \emph{black} use random and black perturbation respectively during the insertion metric. The bold values beat the highest values for the respective metric in our original experiment.}
\label{tb:appendix_insertion_metric}
\end{table}

\end{document}